\documentclass{article}

\PassOptionsToPackage{square, numbers, sort&compress}{natbib}
\usepackage[final]{neurips_2025_ml4ps}

\usepackage[utf8]{inputenc} 
\usepackage[T1]{fontenc}    
\usepackage{hyperref}       
\usepackage{url}            
\usepackage{booktabs}       
\usepackage{amsfonts}       
\usepackage{nicefrac}       
\usepackage{microtype}      
\usepackage{xcolor}         
\usepackage{graphicx}

\usepackage{siunitx}

\title{Inverse Autoregressive Flows for Zero Degree Calorimeter fast simulation}

\author{%
  Emilia Majerz \\
  AGH University of Krakow \\
  \texttt{majerz@agh.edu.pl} \\
  \And
  Witold Dzwinel \\
  AGH University of Krakow \\
  \texttt{dzwinel@agh.edu.pl} \\
    \And
  Jacek Kitowski \\
  AGH University of Krakow \\
  \texttt{kito@agh.edu.pl} \\
}

\begin{document}

\maketitle

\begin{abstract}
Physics-based machine learning blends traditional science with modern data-driven techniques. Rather than relying exclusively on empirical data or predefined equations, this methodology embeds domain knowledge directly into the learning process, resulting in models that are both more accurate and robust. We leverage this paradigm to accelerate simulations of the Zero Degree Calorimeter (ZDC) of the ALICE experiment at CERN. Our method introduces a novel loss function and an output variability-based scaling mechanism, which enhance the model’s capability to accurately represent the spatial distribution and morphology of particle showers in detector outputs while mitigating the influence of rare artefacts on the training. Leveraging Normalizing Flows (NFs) in a teacher-student generative framework, we demonstrate that our approach not only outperforms classic data-driven model assimilation but also yields models that are 421 times faster than existing NF implementations in ZDC simulation literature.
\end{abstract}

\section{Introduction}

The storage and processing demands at CERN are rapidly growing~\cite{cerncomputing}, creating the need to replace time- and resource-intensive simulation engines with faster surrogates. The most promising alternatives include generative neural frameworks such as Generative Adversarial Networks (GANs)~\cite{Paganini2017CaloGANS3, krause2024calochallenge2022communitychallenge, giannelli2024caloshowergan, khattak2021fastsimulationhighgranularity}, autoencoders~\cite{Buhmann_2021, krause2024calochallenge2022communitychallenge, liu2024calovqvectorquantizedtwostagegenerative, cresswell2022calomanfastgenerationcalorimeter}, Normalizing Flows (NFs)~\cite{Krause_2023, krause2024calochallenge2022communitychallenge, 10.21468/SciPostPhys.17.2.045, cresswell2022calomanfastgenerationcalorimeter, PhysRevD.107.113004, Diefenbacher_2023, PhysRevD.109.033006, 10.21468/SciPostPhys.18.3.081}, diffusion models~\cite{Mikuni_2022, krause2024calochallenge2022communitychallenge, amram2023denoisingdiffusionmodelsgeometry, Buhmann_2023}, and Flow Matching~\cite{favaro2024calodreamdetectorresponse, krause2024calochallenge2022communitychallenge}. 

The particle transport from the interaction point, where the collision occurs, to the Zero Degree Calorimeter (ZDC)~\cite{Gallio:381433} from the ALICE experiment is particularly long, as the distance between them is \SI{112.5}{\metre}. The simulation involving the GEANT4~\cite{AGOSTINELLI2003250} environment allows for obtaining physically accurate results, but at the cost of a long computation time. To optimise the process, many studies used generative frameworks~\cite{9311504, Dubinski2023MachineLM, kita2024generativediffusionmodelsfast, zdcfastsim, 10.1007/978-3-031-30105-6_22}, mainly on ZDC's neutron sub-detector (ZN).


The traditional procedure for surrogate model training is purely data-driven. On the other hand, simulation of the ZDC responses is based on physical theory and detector setup, creating a perfect use case for Physics-Based Deep Learning~\cite{thuerey2025physicsbaseddeeplearning}, where theory interplays with data to create powerful machine learning (ML) models. Among the reported approaches to ZN modelling, only one can be considered physics-based. During transport, physical processes may cause particle decays, and the particles that reach the detector may be products of them. Thus, simulations of the exact same particle can give significantly different valid results. In~\cite{10.1007/978-3-031-30105-6_22}, the authors addressed the mode collapse issue in GANs, which limits the variability of GAN-generated detector responses. In this work, a regularization term was added to the training loss and scaled by a dataset-derived value, which determined the level of variability in the detector outputs given the same input particle.

NFs are able to model complex probability distributions through a sequence of invertible transformations applied to a simple base distribution. The stochastic aspect of NFs is particularly important in ZN-like simulations, as it inherently captures the variability of detector responses, which makes them a strong choice for generating detector responses based on input particles. Yet, only one research modelled ZN through NFs~\cite{zdcfastsim}. The authors trained a Masked Autoregressive Flow (MAF)~\cite{NIPS2017_6c1da886}-based model that generated high-quality samples, although the inference was slow, as MAFs are fast in likelihood evaluation and slow in sampling. Conversely, Inverse Autoregressive Flows (IAFs)~\cite{kingma2017improvingvariationalinferenceinverse} sample faster at the cost of slow density estimation, which leads to long and difficult training for generative tasks. In the calorimeter modelling area, the teacher-student training allows for obtaining fast IAF student models closely mimicking the behaviour of previously trained MAF teachers~\cite{PhysRevD.107.113004}.

Therefore, we propose to enhance the MAF-IAF training with physics-based features of the modelled detector. We achieve high-fidelity surrogate models, which are significantly faster than the NFs reported before for ZDC. The physics-based part improves the mapping between particular input particle features and detector responses.

\section{Dataset and metrics}

The dataset contains 306,780 samples generated by GEANT4.  Each sample consists of an input particle features vector and a ZN response. The ZN detects light produced by particles passing through its fibres, and the reported values are the numbers of photons detected in each fibre. Here, we treat the responses as 1-channel images of size 44x44. Input vectors contain values indicating particle energy, 3-momentum, position (also in 3 dimensions), mass, charge, and the total number of photons in the detector response. Since this last value is calculated from the result image, it is intended to be provided during the inference by an additional ML model trained on the remaining particle features. 

The dataset includes 21 different input particle types. Here, we perform experiments on the whole dataset, and also on subsets containing particles such as \textit{neutron} ($n$), \textit{lambda} ($\Lambda$), \textit{k-short} ($K_S^0$), and \textit{sigma+} ($\Sigma^+$) particles, respectively. These particles cover vastly different parts of the whole dataset - 23\%, 3\%, 2\%, and 0.5\%, respectively, and produce responses with different levels of \textit{diversity}. This standard deviation-based feature can be quantified with:
\begin{equation}\label{eq-div}
    f_{div}(c) = \sum_{i,j} \sqrt{\frac{\sum_{t} (x^t_{ij} - \mu_{ij})^2}{|X_c|}}, 
\end{equation}
where $x^t_{ij}$ is the pixel value at coordinates $i, j$ for a detector response $t$, $\mu_{ij}$ is the mean of these pixel values for the unique condition input vector $\mathbf{c}$, and $|X_c|$ is the number of detector responses in the dataset for the input $\mathbf{c}$~\cite{10.1007/978-3-031-30105-6_22}.

The main evaluation metric established in the ZDC literature is a mean 1-Wasserstein distance between distributions of five values (\textit{channels}) computed from reference and generated data. The channels are related to the physical detector setup and determine the number of photons collected from five parts of the detector response, which are then sent for further processing. The first value is the sum of outputs of every second fibre (checkerboard-like pattern), and the remaining fibres are divided into four groups arranged in a 2x2 grid (top-left, top-right, bottom-left, bottom-right), and their outputs are also summed.

While the Wasserstein distance captures the overall distribution of the responses, it does not reflect how accurately the model maps input vectors to particular outputs; we propose two additional metrics to quantify this dependence. For each unique input vector $\mathbf{c}$, we compute average channel values from reference and generated data, and calculate Mean Absolute Error (MAE) between them. We report the average of such values ($MAE_c$) and the weighted average, with weights indicating the number of occurrences of each unique input in the dataset ($MAE_{cw}$).



\section{Methodology}

The trained teacher NF models are MAF-based; the transformations are expressed as rational quadratic splines (RQS)~\cite{d1220ead849147929d687c8a42669907}, with parameters learned by Masked Autoencoders for Distribution Estimation (MADE)~\cite{germain2015mademaskedautoencoderdistribution} blocks - the same configuration as in~\cite{zdcfastsim}. The baseline IAF training was modelled after~\cite{PhysRevD.107.113004}. The teacher-student training involves leveraging the respective fast passes of the models, with mappings between data ($x$) and latent noise ($z$) in the directions: $x \rightarrow z \rightarrow x'$ (data loop) and $z \rightarrow x' \rightarrow z'$ (latent loop). The differences between all intermediate outputs from the teacher and the student are minimised, as well as the differences between outputs from respective MADE blocks, to ensure close teacher-student alignment. This setup serves as the baseline for our experiments.

Since the optimisation is driven by Mean Squared Errors (MSE) between student and teacher outputs, it may cause the student to focus too much on modelling some nuance values, and compromise the global structure of the responses. Thus, to guide the optimisation in the correct direction, we propose an additional physics-based loss term based on channel values computed from student outputs in the data loop. Its goal is to help with the modelling of the position and shape of the particle shower observed in the response. As the images undergo preprocessing including noise addition, normalisation by a sum of the pixel values, and logit transformation, we compute the channel values on student responses with sigmoid applied, followed by the normalisation, as on:
\begin{equation}\label{channel-loss}
    \mathcal{L}_{channel} = \frac{1}{n} \sum_{k=1}^n \sum_{i=1}^m (w_i^k - \hat{w}_i^k)^2,
\end{equation}
where $w_i^k$ is the \textit{i}-th channel value for the $k$-th sample, $n$ is the number of samples in the dataset, $m$ is the number of channels, and dash indicates predicted values.  While such responses are not the final samples (they lack rescaling by photon sum and denoising), the channel values computed from them already capture their global structures.

The high level of \textit{diversity} of some of the samples makes it harder for the models to capture their underlying distribution. However, many such events are rare and under-represented in the dataset, and their modelling is not crucial for the whole task. We propose combining the loss with a \textit{diversity}-based scaler, which puts less focus on rare artefacts, while promoting the optimisation towards the better-represented samples:
\begin{equation}\label{diversity-scaler-weighted}
    f_{div\_inv\_w}(c) = [\frac{1}{f_{div}(c)}]_{norm} \times |X_c| + \epsilon .
\end{equation}
The value is computed for each unique input vector; it is based on the inverse \textit{diversity} value normalised to range $[0; 1]$ and reweighted by the number of occurrences of each vector $\mathbf{c}$ in the dataset. We also add a small constant $\epsilon$ to the values to prevent multiplying the loss by 0, which has the effect of removing samples from training. The reweighting ensures that the highly diverse vectors common in the dataset are not overlooked during the optimisation, and only rare artefacts are less important for the process. We compute these values from the training dataset, providing additional external knowledge during the training, which focuses on the assimilation of the student into the teacher. 


\section{Results}

We divided the dataset into training, validation, and test sets (70:10:20), and evaluated four setups: the baseline (with MSE-based assimilation - bs), the baseline with inverted \textit{diversity}-based weighting (bs+div), the baseline with channel loss (bs+ch), and the baseline with channel loss where both MSE and channel loss are weighted (bs+ch+div). We present the results in Table~\ref{tab:results}.

\begin{table}
  \caption{Comparison of IAF student performances in baseline and physics-based settings. The best results for each particle group and each metric are in \textbf{bold}. For reference, teacher results are presented in the first row.}
  \label{tab:results}
  \scriptsize
  \centering
  \begin{tabular}{llllll}
    \toprule
    & $n$ & $\Lambda$ & $K_S^0$ & $\Sigma^+$ & all \\
    \midrule
    teacher WS & $3.56 \pm 0.07$ & $16.00 \pm 0.26$ & $10.94 \pm 0.23$ & $31.52 \pm 0.64$ & $1.74 \pm 0.02$ \\
    teacher $MAE_{c}$ & $7.41 \pm 0.12$ & $8.88 \pm 0.25$ & $9.26 \pm 0.15$ & $12.93 \pm 0.46$ & $8.89 \pm 0.04$ \\
    teacher $MAE_{cw}$ & $6.11 \pm 0.06$ & $9.39 \pm 0.15$ & $11.41 \pm 0.20$ & $23.83 \pm 0.55$ & $5.17 \pm 0.03$ \\
    \midrule
    bs WS & $\mathbf{3.37 \pm 0.08}$ & $16.42 \pm 0.42$ & $10.55 \pm 0.18$ & $33.53 \pm 0.31$ & $1.73 \pm 0.02$ \\
    bs $MAE_{c}$ & $7.65 \pm 0.07$ & $9.57 \pm 0.21$ & $9.40 \pm 0.14$ & $13.84 \pm 0.19$ & $9.16 \pm 0.04$ \\
    bs $MAE_{cw}$ & $6.45 \pm 0.08$ & $11.00 \pm 0.17$ & $11.87 \pm 0.16$ & $25.38 \pm 0.34$ & $5.57 \pm 0.03$ \\
    \midrule
    bs+div WS & $3.45 \pm 0.05$ & $\mathbf{16.32 \pm 0.14}$ & $11.02 \pm 0.25$ & $33.60 \pm 0.16$ & $\mathbf{1.62 \pm 0.02}$ \\
    bs+div $MAE_{c}$ & $7.64 \pm 0.08$ & $9.02 \pm 0.20$ & $9.63 \pm 0.27$ & $13.83 \pm 0.17$ & $9.05 \pm 0.04$ \\
    bs+div $MAE_{cw}$ & $6.40 \pm 0.05$ & $10.56 \pm 0.24$ & $12.29 \pm 0.13$ & $25.75 \pm 0.33$ & $5.40 \pm 0.04$ \\
    \midrule
    bs+ch WS & $4.24 \pm 0.05$ & $16.61 \pm 0.12$ & $10.86 \pm 0.21$ & $\mathbf{31.25 \pm 0.49}$ & $2.00 \pm 0.02$ \\
    bs+ch $MAE_{c}$ & $7.49 \pm 0.05$ & $8.66 \pm 0.25$ & $9.30 \pm 0.08$ & $\mathbf{11.13 \pm 0.12}$ & $8.89 \pm 0.02$ \\
    bs+ch $MAE_{cw}$ & $6.32 \pm 0.07$ & $9.22 \pm 0.20$ & $11.24 \pm 0.25$ & $\mathbf{18.47 \pm 0.66}$ & $5.17 \pm 0.02$ \\
    \midrule
    bs+ch+div WS & $4.19 \pm 0.07$ & $16.49 \pm 0.38$ & $\mathbf{10.49 \pm 0.16}$ & $31.64 \pm 0.32$ & $1.71 \pm 0.02$ \\
    bs+ch+div $MAE_{c}$ & $\mathbf{7.46 \pm 0.07}$ & $\mathbf{8.18 \pm 0.09}$ & $\mathbf{9.03 \pm 0.08}$ & $11.98 \pm 0.22$ & $\mathbf{8.77 \pm 0.06}$ \\
    bs+ch+div $MAE_{cw}$ & $\mathbf{6.20 \pm 0.04}$ & $\mathbf{8.88 \pm 0.10}$ & $\mathbf{10.75 \pm 0.20}$ & $19.14 \pm 0.84$ & $\mathbf{5.00 \pm 0.02}$ \\
    \bottomrule
  \end{tabular}
\end{table}

In the preparation stage, we trained five separate teachers: for four particle types ($n$, $\Lambda$, $K_S^0$, and $\Sigma^+$) and for the full dataset. Then, we performed the experiments independently for the five teachers and four setups. We report means and standard deviations calculated over five runs of generating responses on the test data. 

The results show a weakness in the use of the Wasserstein score alone. A lower value does not always come with a lower value of metrics measuring how well the model maps input features to a particular output. It indicates that while the global structure is better captured, the physical dependencies are (at least partially) broken. Thus, we propose to move the focus to the metrics measuring physical relevance between inputs and outputs, with additional insights given by a global metric.

\begin{figure}
  \centering
  \includegraphics[width=0.8\linewidth]{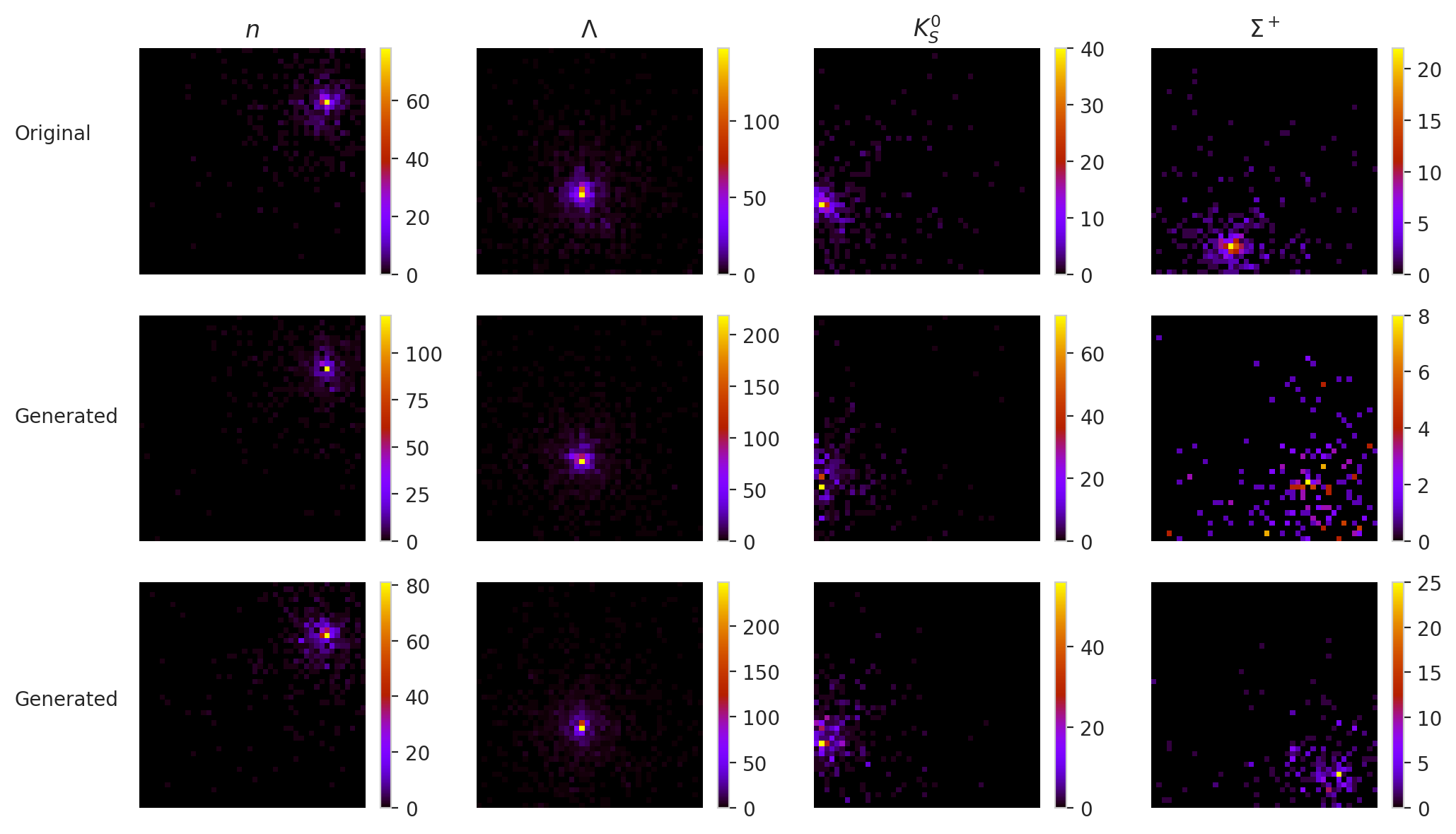}
  \caption{Sample results generated with IAF students. We present two generated samples for each input vector to show that the responses differ between runs.}
  \label{fig:samples}
\end{figure}

The experiments confirmed that the physical dependencies are better captured with the channel loss; in all cases, the ${MAE}_{c}$ and ${MAE}_{cw}$ values are lower when it is applied. While the inverted \textit{diversity}-based weighting alone does not have a consistent impact on the results, combining it with channel loss (also weighted) gave the best results in 4 of 5 cases, and second-best in the one remaining. The Wasserstein values do not deviate much between the models within the same particle group, and we consider models with slightly worse Wasserstein and better MAEs more physically relevant. We present sample results generated with dedicated students trained in the bs+ch+div setup in Fig.~\ref{fig:samples}.

Interestingly, some of the best-performing students exhibit even better performance than their teachers. It is only in the case of models trained with an additional physics-based part, which highlights the usability of dedicated physics-based elements in the training. Moreover, the students are 421 times faster than the teachers - the generating time per sample equals \SI{0.38}{\ms}, which makes a significant advancement to the previously reported \SI{160.0}{\ms} for NFs in ZDC modelling~\cite{zdcfastsim}.

We additionally evaluated metrics that directly assess the accuracy of model predictions regarding the position and shape of the generated showers. For spatial position, we compared the coordinates of the reference and generated shower centres, defined as the energy-weighted centres of mass. To characterise the shape, we used the radius of the circle containing 90\% of the total number of photons in each response. The values were computed for each unique input vector by calculating the differences in the means and variances of these quantities from the reference and generated data, and then aggregated using a weighted average. We analysed both the MAEs and the Root Mean Squared Errors (RMSEs), evaluating the performance of the teacher, bs, and bs+ch+div setups. Accounting for variance is crucial here, as all showers display inherent diversity, which can strongly influence low-level statistics such as a shower position - often to a much greater extent than for the higher-level channel values.


For the four individual particle types, the results were inconclusive; the differences were minor, and different setups were favoured depending on the metric. In comparison, the findings presented in Table~\ref{tab:results} clearly indicate the advantage of the physics-based approach. It suggests that excelling separately in detailed statistics, such as shower position and shape, is not essential for producing better higher-level results, e.g., channel values. Additionally, the limited size of the per-particle sub-datasets may have prevented these low-level metrics from accurately reflecting performance differences across student optimisation approaches. Nonetheless, we provide results for the full dataset in Table~\ref{tab:center-radi}. In this comprehensive case, the student models consistently favour the physics-based setup in all comparisons. Due to the much larger sample size, these results are more reliable than those obtained for individual particle types.


\begin{table}
  \caption{Comparison of the performances of IAF student baseline (bs) and physics-based (bs+ch+div) models regarding the correctness of the modelling of shower positions (centre and centre variance errors) and shapes (radius and radius variance errors) in the "full dataset" (all) case. The best results in each column are in \textbf{bold}. For reference, teacher results are presented in the first row.}
  \label{tab:center-radi}
  \small
  \centering
  \resizebox{\columnwidth}{!}{%
  \begin{tabular}{lllllllll}
    \toprule
    & \multicolumn{2}{l}{centre error} & \multicolumn{2}{l}{centre variance error} & \multicolumn{2}{l}{radius error} & \multicolumn{2}{l}{radius variance error}\\
    & MAE & RMSE & MAE & RMSE & MAE & RMSE & MAE & RMSE \\
    \midrule
    teacher & $1.60 \pm 0.01$ & $5.84 \pm 0.02$ & $11.25 \pm 0.10$ &  $45.68 \pm 0.54$ & $2.22 \pm 0.02$ & $6.62 \pm 0.02$ & $11.29 \pm 0.19$ & $52.66 \pm 1.56$ \\
    bs & $1.64 \pm 0.01$ & $5.86 \pm 0.02$ & $11.16 \pm 0.13$ & $46.90 \pm 0.64$ & $2.34 \pm 0.02$ & $6.69 \pm 0.02$ & $11.27 \pm 0.17$ & $55.09 \pm 1.42$ \\
    bs+ch+div & $\mathbf{1.63 \pm 0.01}$ & $\mathbf{5.83 \pm 0.02}$ & $\mathbf{10.86 \pm 0.13}$ & $\mathbf{45.26 \pm 0.22}$ & $\mathbf{2.26 \pm 0.01}$ & $\mathbf{6.65 \pm 0.03}$ & $\mathbf{11.22 \pm 0.41}$ & $\mathbf{52.77 \pm 2.76}$ \\
    \bottomrule
  \end{tabular}
  }
\end{table}


\section{Conclusions}

Here, we introduce two physics-based approaches for obtaining high-fidelity IAF student models for the ZDC fast simulation. We propose an additional loss based on the physical detector setting, which helps in modelling the position and shape of the particle shower present in the detector response. By weighting the loss with the response variability-induced scaler, we lessen the impact of the rare artefacts on the overall training. Our experiments show that minor adjustments improve the modelling of the underlying physical dependencies between input particles and detector outputs, which is crucial in the particle physics research area. Moreover, our models are 421 times faster than the previously reported NFs in ZDC simulation research, establishing them as a strong method for simulating this detector.

\acksection
We would like to thank Professor Jacek Otwinowski from the Institute of Nuclear Physics PAS in Krakow for his support in this work. This work is co-financed and in part supported by the Ministry of Science and Higher Education  (Agreement No. 2023/WK/07) by the program entitled ``PMW'' and by the Ministry funds assigned to AGH University of Krakow. We gratefully acknowledge Polish high-performance computing infrastructure PLGrid (HPC Center: ACK Cyfronet AGH) for providing computer facilities and support within computational grants no. PLG/2024/017264 and PLG/2025/018322.

\bibliography{sample}

@misc{cerncomputing,
    author    = "CERN",
    title     = "Computing | CERN",
    howpublished       = "\url{https://home.cern/science/computing}",
    year      = "2025",
    note      = "[Online; accessed 27-August-2025]"
}

@article{Paganini2017CaloGANS3,
  title = {CaloGAN: Simulating 3D high energy particle showers in multilayer electromagnetic calorimeters with generative adversarial networks},
  author = {Paganini, Michela and de Oliveira, Luke and Nachman, Benjamin},
  journal = {Phys. Rev. D},
  volume = {97},
  issue = {1},
  pages = {014021},
  numpages = {12},
  year = {2018},
  month = {Jan},
  publisher = {American Physical Society},
  doi = {10.1103/PhysRevD.97.014021},
  url = {https://link.aps.org/doi/10.1103/PhysRevD.97.014021}
}

@misc{krause2024calochallenge2022communitychallenge,
      title={CaloChallenge 2022: A Community Challenge for Fast Calorimeter Simulation}, 
      author={Claudius Krause and Michele Faucci Giannelli and Gregor Kasieczka and Benjamin Nachman and Dalila Salamani and David Shih and Anna Zaborowska and Oz Amram and Kerstin Borras and Matthew R. Buckley and Erik Buhmann and Thorsten Buss and Renato Paulo Da Costa Cardoso and Anthony L. Caterini and Nadezda Chernyavskaya and Federico A. G. Corchia and Jesse C. Cresswell and Sascha Diefenbacher and Etienne Dreyer and Vijay Ekambaram and Engin Eren and Florian Ernst and Luigi Favaro and Matteo Franchini and Frank Gaede and Eilam Gross and Shih-Chieh Hsu and Kristina Jaruskova and Benno Käch and Jayant Kalagnanam and Raghav Kansal and Taewoo Kim and Dmitrii Kobylianskii and Anatolii Korol and William Korcari and Dirk Krücker and Katja Krüger and Marco Letizia and Shu Li and Qibin Liu and Xiulong Liu and Gabriel Loaiza-Ganem and Thandikire Madula and Peter McKeown and Isabell-A. Melzer-Pellmann and Vinicius Mikuni and Nam Nguyen and Ayodele Ore and Sofia Palacios Schweitzer and Ian Pang and Kevin Pedro and Tilman Plehn and Witold Pokorski and Huilin Qu and Piyush Raikwar and John A. Raine and Humberto Reyes-Gonzalez and Lorenzo Rinaldi and Brendan Leigh Ross and Moritz A. W. Scham and Simon Schnake and Chase Shimmin and Eli Shlizerman and Nathalie Soybelman and Mudhakar Srivatsa and Kalliopi Tsolaki and Sofia Vallecorsa and Kyongmin Yeo and Rui Zhang},
      year={2024},
      eprint={2410.21611},
      archivePrefix={arXiv},
      primaryClass={physics.ins-det},
      url={https://arxiv.org/abs/2410.21611}, 
}

@article{giannelli2024caloshowergan,
  title={CaloShowerGAN, a generative adversarial network model for fast calorimeter shower simulation},
  author={Giannelli, Michele Faucci and Zhang, Rui},
  journal={The European Physical Journal Plus},
  volume={139},
  number={7},
  pages={597},
  year={2024},
  publisher={Springer}
}

@misc{khattak2021fastsimulationhighgranularity,
      title={Fast Simulation of a High Granularity Calorimeter by Generative Adversarial Networks}, 
      author={Gul Rukh Khattak and Sofia Vallecorsa and Federico Carminati and Gul Muhammad Khan},
      year={2021},
      eprint={2109.07388},
      archivePrefix={arXiv},
      primaryClass={physics.ins-det},
      url={https://arxiv.org/abs/2109.07388}, 
}

@article{Buhmann_2021,
   title={Getting High: High Fidelity Simulation of High Granularity Calorimeters with High Speed},
   volume={5},
   ISSN={2510-2044},
   url={http://dx.doi.org/10.1007/s41781-021-00056-0},
   DOI={10.1007/s41781-021-00056-0},
   number={1},
   journal={Computing and Software for Big Science},
   publisher={Springer Science and Business Media LLC},
   author={Buhmann, Erik and Diefenbacher, Sascha and Eren, Engin and Gaede, Frank and Kasieczka, Gregor and Korol, Anatolii and Krüger, Katja},
   year={2021},
   month=may }

@misc{liu2024calovqvectorquantizedtwostagegenerative,
      title={Calo-VQ: Vector-Quantized Two-Stage Generative Model in Calorimeter Simulation}, 
      author={Qibin Liu and Chase Shimmin and Xiulong Liu and Eli Shlizerman and Shu Li and Shih-Chieh Hsu},
      year={2024},
      eprint={2405.06605},
      archivePrefix={arXiv},
      primaryClass={physics.ins-det},
      url={https://arxiv.org/abs/2405.06605}, 
}

@misc{cresswell2022calomanfastgenerationcalorimeter,
      title={CaloMan: Fast generation of calorimeter showers with density estimation on learned manifolds}, 
      author={Jesse C. Cresswell and Brendan Leigh Ross and Gabriel Loaiza-Ganem and Humberto Reyes-Gonzalez and Marco Letizia and Anthony L. Caterini},
      year={2022},
      eprint={2211.15380},
      archivePrefix={arXiv},
      primaryClass={hep-ph},
      url={https://arxiv.org/abs/2211.15380}, 
}

@article{Krause_2023,
   title={Fast and accurate simulations of calorimeter showers with normalizing flows},
   volume={107},
   ISSN={2470-0029},
   url={http://dx.doi.org/10.1103/PhysRevD.107.113003},
   DOI={10.1103/physrevd.107.113003},
   number={11},
   journal={Physical Review D},
   publisher={American Physical Society (APS)},
   author={Krause, Claudius and Shih, David},
   year={2023},
   month=jun }

@Article{10.21468/SciPostPhys.17.2.045,
	title={{Towards a data-driven model of hadronization using normalizing flows}},
	author={Christian Bierlich and Phil Ilten and Tony Menzo and Stephen Mrenna and Manuel Szewc and Michael K. Wilkinson and Ahmed Youssef and Jure Zupan},
	journal={SciPost Phys.},
	volume={17},
	pages={045},
	year={2024},
	publisher={SciPost},
	doi={10.21468/SciPostPhys.17.2.045},
	url={https://scipost.org/10.21468/SciPostPhys.17.2.045},
}

@article{PhysRevD.107.113004,
  title = {Accelerating accurate simulations of calorimeter showers with normalizing flows and probability density distillation},
  author = {Krause, Claudius and Shih, David},
  journal = {Phys. Rev. D},
  volume = {107},
  issue = {11},
  pages = {113004},
  numpages = {19},
  year = {2023},
  month = {Jun},
  publisher = {American Physical Society},
  doi = {10.1103/PhysRevD.107.113004},
  url = {https://link.aps.org/doi/10.1103/PhysRevD.107.113004}
}

@article{Diefenbacher_2023,
doi = {10.1088/1748-0221/18/10/P10017},
url = {https://dx.doi.org/10.1088/1748-0221/18/10/P10017},
year = {2023},
month = {oct},
publisher = {IOP Publishing},
volume = {18},
number = {10},
pages = {P10017},
author = {Diefenbacher, Sascha and Eren, Engin and Gaede, Frank and Kasieczka, Gregor and Krause, Claudius and Shekhzadeh, Imahn and Shih, David},
title = {L2LFlows: generating high-fidelity 3D calorimeter images},
journal = {Journal of Instrumentation}
}

@article{PhysRevD.109.033006,
  title = {Inductive simulation of calorimeter showers with normalizing flows},
  author = {Buckley, Matthew R. and Pang, Ian and Shih, David and Krause, Claudius},
  journal = {Phys. Rev. D},
  volume = {109},
  issue = {3},
  pages = {033006},
  numpages = {19},
  year = {2024},
  month = {Feb},
  publisher = {American Physical Society},
  doi = {10.1103/PhysRevD.109.033006},
  url = {https://link.aps.org/doi/10.1103/PhysRevD.109.033006}
}

@Article{10.21468/SciPostPhys.18.3.081,
	title={{Normalizing flows for high-dimensional detector simulations}},
	author={Florian Ernst and Luigi Favaro and Claudius Krause and Tilman Plehn and David Shih},
	journal={SciPost Phys.},
	volume={18},
	pages={081},
	year={2025},
	publisher={SciPost},
	doi={10.21468/SciPostPhys.18.3.081},
	url={https://scipost.org/10.21468/SciPostPhys.18.3.081},
}

@article{Mikuni_2022,
   title={Score-based generative models for calorimeter shower simulation},
   volume={106},
   ISSN={2470-0029},
   url={http://dx.doi.org/10.1103/PhysRevD.106.092009},
   DOI={10.1103/physrevd.106.092009},
   number={9},
   journal={Physical Review D},
   publisher={American Physical Society (APS)},
   author={Mikuni, Vinicius and Nachman, Benjamin},
   year={2022},
   month=nov }

@misc{favaro2024calodreamdetectorresponse,
      title={CaloDREAM -- Detector Response Emulation via Attentive flow Matching}, 
      author={Luigi Favaro and Ayodele Ore and Sofia Palacios Schweitzer and Tilman Plehn},
      year={2024},
      eprint={2405.09629},
      archivePrefix={arXiv},
      primaryClass={hep-ph},
      url={https://arxiv.org/abs/2405.09629}, 
}

@article{Buhmann_2023,
   title={CaloClouds: fast geometry-independent highly-granular calorimeter simulation},
   volume={18},
   ISSN={1748-0221},
   url={http://dx.doi.org/10.1088/1748-0221/18/11/P11025},
   DOI={10.1088/1748-0221/18/11/p11025},
   number={11},
   journal={Journal of Instrumentation},
   publisher={IOP Publishing},
   author={Buhmann, Erik and Diefenbacher, Sascha and Eren, Engin and Gaede, Frank and Kasicezka, Gregor and Korol, Anatolii and Korcari, William and Krüger, Katja and McKeown, Peter},
   year={2023},
   month=nov, pages={P11025} }

@misc{amram2023denoisingdiffusionmodelsgeometry,
      title={Denoising diffusion models with geometry adaptation for high fidelity calorimeter simulation}, 
      author={Oz Amram and Kevin Pedro},
      year={2023},
      eprint={2308.03876},
      archivePrefix={arXiv},
      primaryClass={physics.ins-det},
      url={https://arxiv.org/abs/2308.03876}, 
}

@book{Gallio:381433,
      author        = "Gallio, M and Klempt, W and Leistam, L and De Groot, J and
                       Schukraft, Jürgen",
      collaboration = "ALICE",
      title         = "{ALICE Zero-Degree Calorimeter (ZDC): Technical Design
                       Report}",
      publisher     = "CERN",
      address       = "Geneva",
      series        = "Technical design report. ALICE",
      year          = "1999",
      url           = "https://cds.cern.ch/record/381433",
}

@article{AGOSTINELLI2003250,
    author = "Agostinelli, S. and others",
    collaboration = "GEANT4",
    title = "{GEANT4 - A Simulation Toolkit}",
    reportNumber = "SLAC-PUB-9350, FERMILAB-PUB-03-339, CERN-IT-2002-003",
    doi = "10.1016/S0168-9002(03)01368-8",
    journal = "Nuclear Instruments and Methods in Physics Research",
    volume = "506",
    pages = "250--303",
    year = "2003"
}

@ARTICLE{9311504,
  author={Deja, Kamil and Dubiński, Jan and Nowak, Piotr and Wenzel, Sandro and Spurek, Przemysław and Trzcinski, Tomasz},
  journal={IEEE Access}, 
  title={End-to-End Sinkhorn Autoencoder With Noise Generator}, 
  year={2021},
  volume={9},
  number={},
  pages={7211-7219},
  doi={10.1109/ACCESS.2020.3048622}}

@misc{Dubinski2023MachineLM,
      title={Machine Learning methods for simulating particle response in the Zero Degree Calorimeter at the ALICE experiment, CERN}, 
      author={Jan Dubiński and Kamil Deja and Sandro Wenzel and Przemysław Rokita and Tomasz Trzciński},
      year={2023},
      eprint={2306.13606},
      archivePrefix={arXiv},
      primaryClass={cs.CV},
      url={https://arxiv.org/abs/2306.13606}, 
}

@misc{kita2024generativediffusionmodelsfast,
      title={Generative Diffusion Models for Fast Simulations of Particle Collisions at CERN}, 
      author={Mikołaj Kita and Jan Dubiński and Przemysław Rokita and Kamil Deja},
      year={2024},
      eprint={2406.03233},
      archivePrefix={arXiv},
      primaryClass={physics.data-an},
      url={https://arxiv.org/abs/2406.03233}, 
}

@InProceedings{10.1007/978-3-031-30105-6_22,
author="Dubi{\'{n}}ski, Jan
and Deja, Kamil
and Wenzel, Sandro
and Rokita, Przemys{\l}aw
and Trzcinski, Tomasz",
editor="Tanveer, Mohammad
and Agarwal, Sonali
and Ozawa, Seiichi
and Ekbal, Asif
and Jatowt, Adam",
title="Selectively Increasing the Diversity of GAN-Generated Samples",
booktitle="Neural Information Processing",
year="2023",
publisher="Springer International Publishing",
address="Cham",
pages="260--270",
isbn="978-3-031-30105-6"
}

@Article{zdcfastsim,
author={Wojnar, Maksymilian
and Majerz, Emilia
and Dzwinel, Witold},
title={Fast Simulation of the Zero Degree Calorimeter Responses with Generative Neural Networks},
journal={Computing and Software for Big Science},
year={2025},
month={Jan},
day={27},
volume={9},
number={1},
pages={1},
issn={2510-2044},
doi={10.1007/s41781-025-00130-x},
url={https://doi.org/10.1007/s41781-025-00130-x}
}

@misc{thuerey2025physicsbaseddeeplearning,
      title={Physics-based Deep Learning}, 
      author={N. Thuerey and B. Holzschuh and P. Holl and G. Kohl and M. Lino and Q. Liu and P. Schnell and F. Trost},
      year={2025},
      eprint={2109.05237},
      archivePrefix={arXiv},
      primaryClass={cs.LG},
      url={https://arxiv.org/abs/2109.05237}, 
}

@inproceedings{NIPS2017_6c1da886,
 author = {Papamakarios, George and Pavlakou, Theo and Murray, Iain},
 booktitle = {Advances in Neural Information Processing Systems},
 editor = {I. Guyon and U. Von Luxburg and S. Bengio and H. Wallach and R. Fergus and S. Vishwanathan and R. Garnett},
 pages = {},
 publisher = {Curran Associates, Inc.},
 title = {Masked Autoregressive Flow for Density Estimation},
 url = {https://proceedings.neurips.cc/paper_files/paper/2017/file/6c1da886822c67822bcf3679d04369fa-Paper.pdf},
 volume = {30},
 year = {2017}
}

@misc{kingma2017improvingvariationalinferenceinverse,
      title={Improving Variational Inference with Inverse Autoregressive Flow}, 
      author={Diederik P. Kingma and Tim Salimans and Rafal Jozefowicz and Xi Chen and Ilya Sutskever and Max Welling},
      year={2017},
      eprint={1606.04934},
      archivePrefix={arXiv},
      primaryClass={cs.LG},
      url={https://arxiv.org/abs/1606.04934}, 
}

@misc{germain2015mademaskedautoencoderdistribution,
      title={MADE: Masked Autoencoder for Distribution Estimation}, 
      author={Mathieu Germain and Karol Gregor and Iain Murray and Hugo Larochelle},
      year={2015},
      eprint={1502.03509},
      archivePrefix={arXiv},
      primaryClass={cs.LG},
      url={https://arxiv.org/abs/1502.03509}, 
}

@inproceedings{d1220ead849147929d687c8a42669907,
title = "Neural Spline Flows",
author = "Conor Durkan and Artur Bekasovs and Iain Murray and George Papamakarios",
year = "2019",
month = dec,
day = "14",
language = "English",
volume = "32",
series = "Advances in Neural Information Processing Systems",
publisher = "Neural Information Processing Systems Foundation, Inc",
pages = "7511--7522",
booktitle = "Advances in Neural Information Processing Systems 32 (NeurIPS 2019)",
note = "33rd Conference on Neural Information Processing Systems, NeurIPS 2019 ; Conference date: 08-12-2019 Through 14-12-2019",
url = "https://neurips.cc/",
}
\end{document}